\documentclass{article} 
\usepackage[preprint]{iclr2024_conference}

\usepackage[utf8]{inputenc} 
\usepackage[T1]{fontenc}    
\usepackage{hyperref}       
\usepackage{url}            
\usepackage{booktabs}       
\usepackage{amsfonts}       
\usepackage{nicefrac}       
\usepackage{microtype}      
\usepackage{xcolor}         

\usepackage{algorithm}
\usepackage{algpseudocode}

\usepackage{booktabs}
\usepackage{multirow}
\usepackage{siunitx}
\usepackage{amssymb}
\usepackage{booktabs}       
\usepackage{amsfonts}       
\usepackage{nicefrac}       
\usepackage{microtype}      
\usepackage{amsmath}
\usepackage{graphicx}
\usepackage{multirow}
\usepackage{booktabs}
\usepackage{wrapfig}

\title{DoseGNN: Improving the Performance of Deep Learning Models in Adaptive Dose-Volume Histogram Prediction through Graph Neural Networks.}

\author{%
  Zehao Dong\textsuperscript{1}  Yixin Chen\textsuperscript{1} Tianyu Zhao\textsuperscript{2} \\
  \textsuperscript{1} Department of Computer Science $ \&$ Engineering, Washington University in St. Louis \\
  \textsuperscript{2} Department of Radiation Oncology, Washington University School of Medicine in St. Louis \\
  \texttt{\{zehao.dong,tzhao\}@wustl.edu} \&
  \texttt{chen@cse.wustl.edu}
}
\begin{document}

\maketitle

\begin{abstract}
   Dose-Volume Histogram (DVH) prediction is fundamental in radiation therapy that facilitate treatment planning, dose evaluation, plan comparison and etc. It helps to increase the ability to deliver precise and effective radiation treatments while managing potential toxicities to healthy tissues as needed to reduce the risk of complications. This paper extends recently disclosed research findings \cite{ dnog2023dosegnn} presented on AAPM (AAPM 65th Annual Meeting $\&$ Exhibition) and includes necessary technique details. The objective is to design efficient deep learning models for DVH prediction on general radiotherapy platform equipped with high performance CBCT system, where input CT images and target dose images to predict may have different origins, spacing and sizes. Deep learning models widely-adopted in DVH prediction task are evaluated on the novel radiotherapy platform, and graph neural networks (GNNs) are shown to be the ideal architecture to construct a plug-and-play framework to improve predictive performance of base deep learning models in the adaptive setting. 
   
\end{abstract}
\section{Introduction}
\label{intro}
Intensity-modulated radiation therapy (IMRT) enhances the precision of high-radiation-dose regions conforming to the planning target volume (PTV) while minimizing radiation exposure to each organ at risk (OAR) \cite{tham2009treatment}. Dose-Volume Histogram (DVH) prediction aids in IMRT from different perspectives, including dose evaluation, plan comparison, adaptive replanning, and etc ~\cite{palta2003intensity}. In the context of radiotherapy, 3D dose prediction \cite{dong2020deep, shiraishi2016knowledge} estimates the three-dimensional distribution of radiation doses within a patient's anatomy to ensure that prescribed dose is delivered accurately to the target while minimizing the dose to surrounding healthy tissues. Once the 3D dose distribution is calculated, a (DVH) is generated, providing a comprehensive view of the dose distribution in relation to the volumes of interest. 

The CBCT (Cone Beam Computed
Tomography) system has been widely used to provide a three-dimensional imaging representation of the patient's anatomy in guiding the delivery of radiation dose. Various deep learning models \cite{cagni2017knowledge,sumida2020convolution} are developed to predict what is likely achievable for patients' DVHs based on these CBCT/CT images. Typically, these models are constructed upon recent advancements in the field of computer vision, such as 3D U-net ~\cite{cciccek20163d} and Vision Transformer ~\cite{han2022survey}.

In many radiotherapy platform equipped with high performance CBCT system \cite{dong2023performance}, some training plans contain mismatched input CT images and target dose images. That is, the 3D CT images and 3D dose image to predict can have different origins, sizes, and spacings. One intuitive heuristic is to generate a 3D CT image by greedily searching a point/pixel in the original 3D CT image for each point/pixel in the 3D dose image based on the geometry distance, then the image-processing deep learning models can be used for the prediction task. However, the CT images after the greedy conversion are distorted and will fail to capture the patient's anatomy accurately. In contrast, another heuristic is to directly apply image-processing deep learning models on the origin CT images, and then take the average of learnt representations of k-nearest points/pixels in CT images to each point/pixel in the dose image for predicting it's corresponding dose value. However, this heuristic is over-smoothed and fail to capture the geometry variance. 

To address aforementioned challenges, we introduce DoseGNN, a plug-and-play architecture based on graph neural networks (GNNs). GNNs \cite{kipf2016semi, Hamilton2017,xu2018how, Velickovic2018GraphAN, dong2022pace, you2018graphrnn, scarselli2008graph, duvenaud2015convolutional,gilmer2017neural, zhang2018end, dong2022cktgnn} have revolutionized the field of representation learning over graphs and is widely used in clinical applications including drug synergy prediction ~\citep{hopkins2008network, dong2023interpreting, podolsky2011combination}, Alzheimer's disease (AD) detection ~\citep{song2019graph, qin2022aiding, 10.21203/rs.3.rs-3576068/v1} and cancer subtype classification ~\citep{lu2003cancer}. Specifically, we formulate the 3D dose prediction problem as a semi-supervised graph regression problem, where the graph consists of points/pixels in the input CT images and target dose images. CT nodes (i.e. nodes associated with points/pixels in the CT images) and dose nodes (i.e. nodes associated with points/pixels in the dose image) are connected if their geometric distance is smaller than a given threshold. Then, the image-processing deep learning models are used as base model to extract features of CT nodes, and a GNN is used to predict dose values of dose nodes.

DoseGNN is evaluated on 20 plans created by certified medical dosimetrists, with 15 used for training and the remaining 5 used for performance testing. Four deep learning models are used as the image-processing base models: Vision Transformer (ViT) ~\cite{han2022survey}, 3D U-net CNN ~\cite{cciccek20163d}, vanilla multilayer perceptron MLP \cite{cybenko1989approximation} and support vector machine SVM \cite{suthaharan2016support}. Experimental results indicate that the porposed plug-and-play framework DoseGNN significantly and consistantly improves the predictive performance in adaptive 3D dose prediction.


\section{Methodology}
\label{metho}
\begin{figure}[t]
\begin{center}
\centerline{\includegraphics[width=0.85\textwidth]{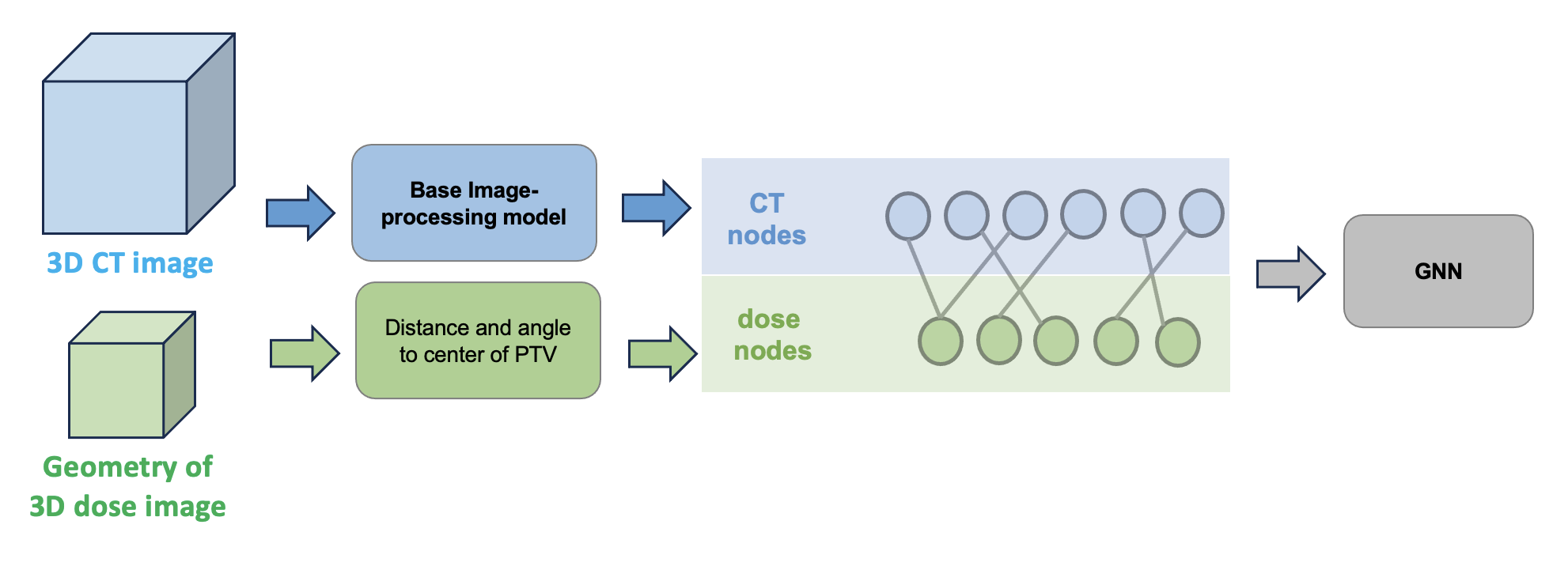}}
\caption{Illustration of the proposed DoseGNN framework. DoseGNN is in a plug-and-play fashion to be integrated arbitrary base image-processing deep learning models. It first extracts geometric features in 3D CT images as the representation of CT nodes, then a GNN is used to predict the target 3D dose image.}
\label{fig:fig1}
\vskip -0.3in
\end{center}
\end{figure}

In this section, we introduce the proposed DoseGNN framework. Figure ~\ref{fig:fig1} illustrates the architecture. DoseGNN is a plug-and-play framework that can be combined with various base image-processing deep learning models. 

\begin{figure}[t]
\begin{center}
\centerline{\includegraphics[width=0.99\textwidth]{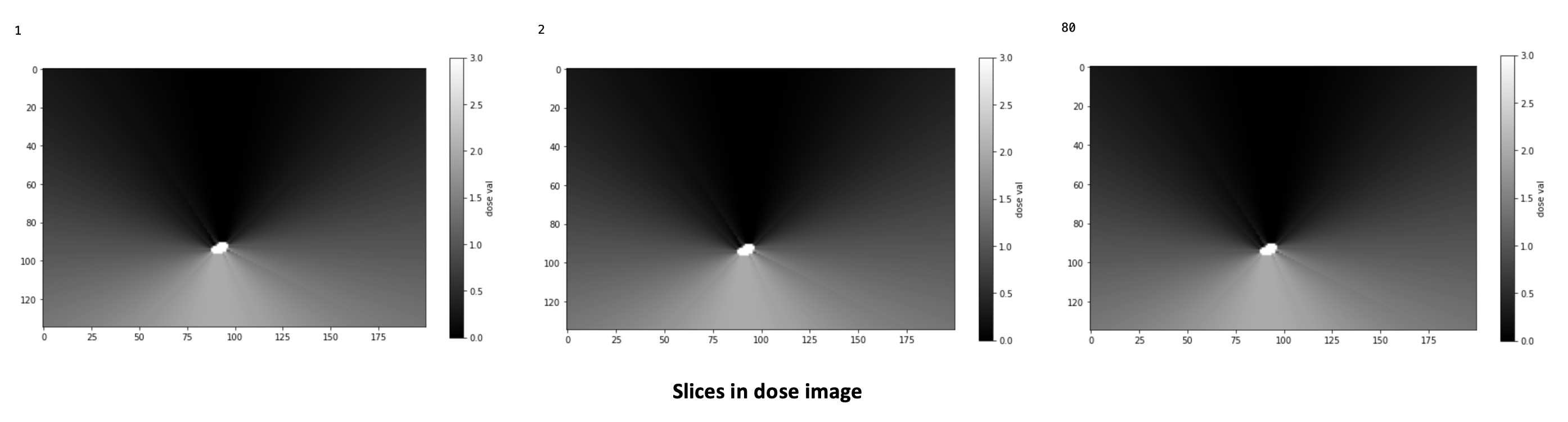}}
\caption{Visualization of slices in 3D dose image.}
\label{fig:fig2}
\vskip -0.3in
\end{center}
\end{figure}

The core idea is to formulate a bipartite graph to modulate the geometry relation of pixels in the input CT images and the target dose images. Specifically, 
\begin{itemize}
    \item For each node in the bipartite graph associated to a point/pixel in the CT image, the base image-processing model is used to extract a vector representation of un-distorted geometric information, which accurately captures patient's anatomy. Thus, the learnt vectorial representation is then used as the node embeddings of CT node.
    \item For each node in the bipartite graph associated to a point/pixel in the dose image, it takes two features: distance and angle to the center of the PTV. As Figure ~\ref{fig:fig2} illustrates, these two features well captures the shape of the dose image distribution. Then, DoseGNN takes the positional encoding framework to generate node embeddings of dose nodes, whose dimension is set to be the same as the embeddings of CT nodes.
    \item In the bipartite graph, a CT node is connected with a dose node if and only if the geometry distance of these two pixels are bounded by a given threshold.
\end{itemize}

Following this pipeline, the 3D dose prediction problem is converted as a semi-supervised graph regression problem. Consequently, a GNN is used to predict the dose value of each dose node in the bipartite graph.


\section{Experiments}
\label{exper}
In this section, we evaluate the effectiveness of proposed DoseGNN framework based on various based image-processing deep learning models. In total, 20 treatment plans created by certified medical dosimetrists in the experiments, with 15 used for training models and the remaining 5 used for performance testing. 

\subsection{Predictive performance}
\label{subsec:datasets}

We use root mean square error (RMSE) of predicted dose and true dose as the evaluation metric in the experiment to evaluate how close the deep learning model can approximate the true dose information. 

\begin{figure}[h]
\begin{center}
\centerline{\includegraphics[width=0.8\textwidth]{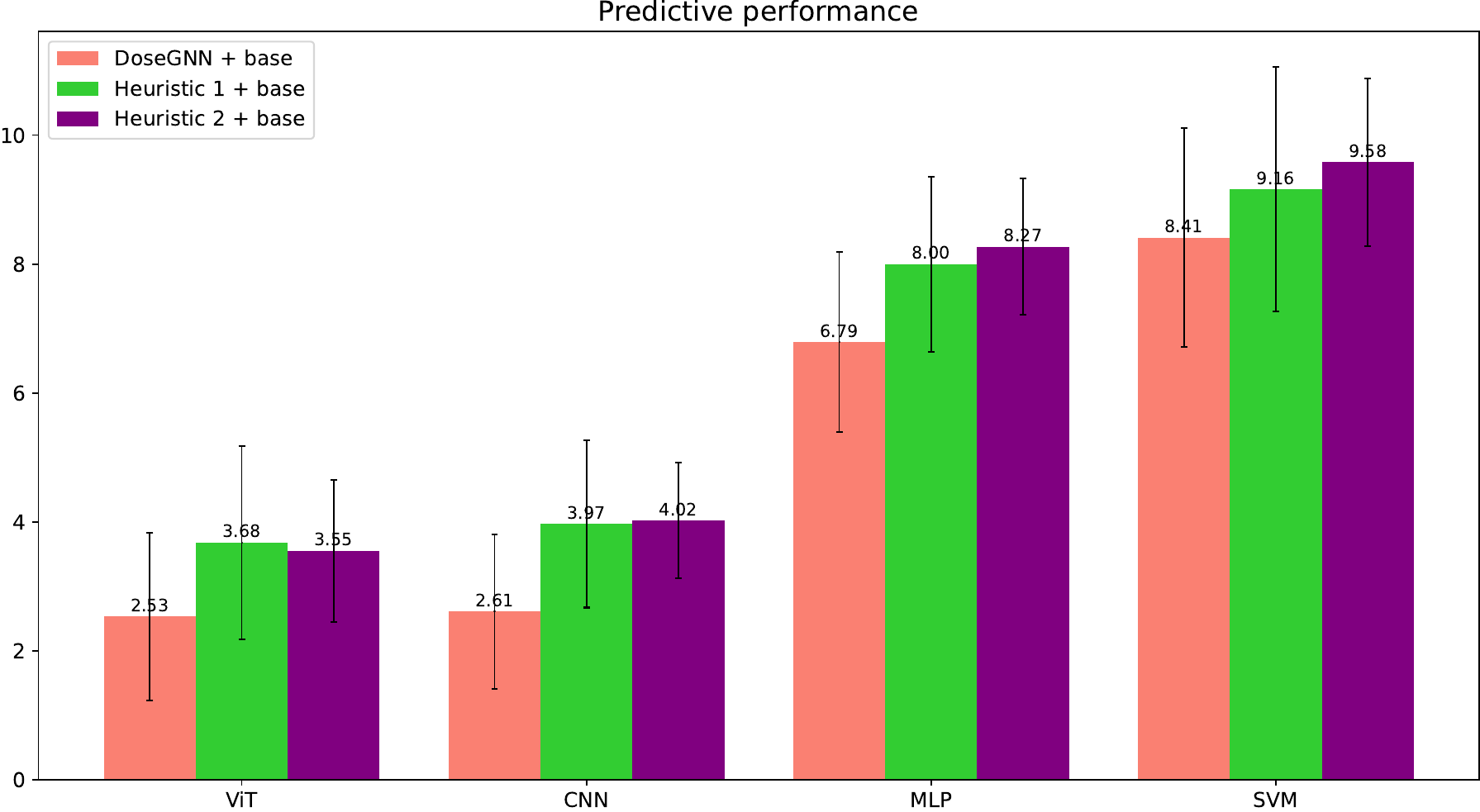}}
\caption{Comparison of 3D dose prediction accuracy}
\label{fig:fig3}
\vskip -0.3in
\end{center}
\end{figure}

Two heuristics are used as baseline method to deal with CT images and dose images of different origins, sizes, and spacings.
\begin{itemize}
    \item Heuristic 1 firstly generates a distorted 3D CT image by greedily searching the nearst point/pixel in the original 3D CT image to each point/pixel in the 3D dose image in terms of the geometry distance, then the image-processing deep learning models is used as a predictor.
    \item Heuristic 2 directly applies image-processing deep learning models on the origin CT images, and then take the average of learnt representations of k-nearest points/pixels in CT images to each point/pixel in the dose image as the final embedding, which is feed to a MLP to predict the corresponding dose value.
\end{itemize}

Figure ~\ref{fig:fig3} presents the results. We find that DoseGNN significantly and consistently improves the baseline heuristics. In addition, DoseGNN with a based of ViT achieves the state-of-the-art performance.

\subsection{Visualization of CDVH}

In the experiment, we visualize the CDVH (cumulative dose volume histogram) of predicted 3D dose from different method. In the experiment, DoseGNN indicates the DoseGNN with a base model of ViT, which beats DoseGNN with other base models. For other base models like ViT and MLP, we visualize the best predictive results the base model can achieve with different heuristics.

\begin{figure}[h]
\begin{center}
\centerline{\includegraphics[width=0.99\textwidth]{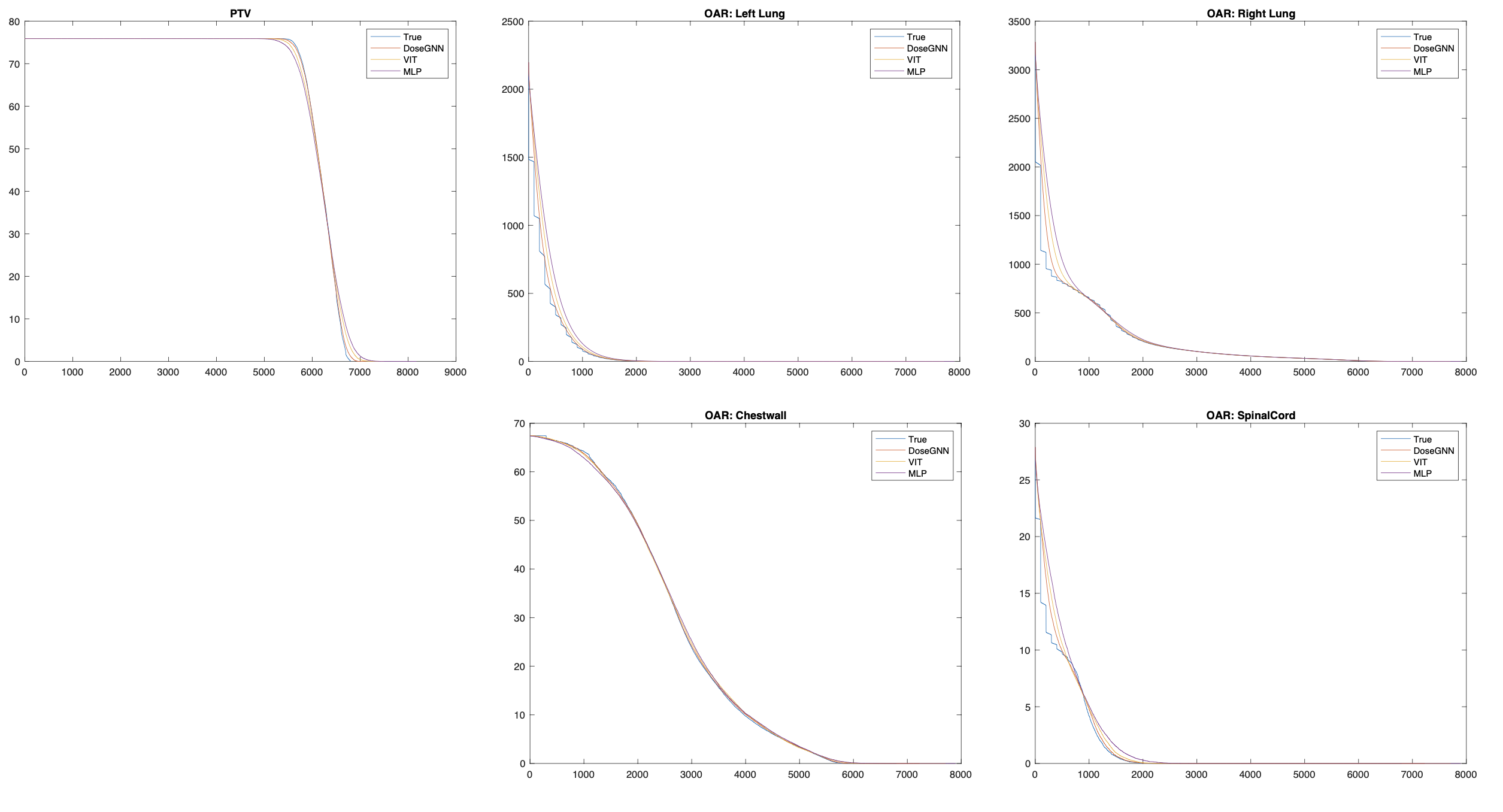}}
\caption{of CDVH of true doses and predicted doses from different deep learning models. }
\label{fig:fig4}
\vskip -0.3in
\end{center}
\end{figure}

In Figure ~\ref{fig:fig4} demonstrates that DoseGNN has the best ability to capture the shape of the CDVH curve of true dose values.

\section{Conclusion}
\label{convlu}
In this paper, we propose a plug-and-play framework, DoseGNN, to enhance the predictive performance of base image-processing deep learning models in adaptive 3D dose prediction tasks. Experimental results demonstrate that DoseGNN significantly and consistently improve base deep learning models, and can be applied in radiotherapy plans with different number of OARs and CT slices.

\newpage

\bibliography{iclr2024_conference}
\bibliographystyle{iclr2024_conference}


\end{document}